\documentclass[conference]{IEEEtran}
\IEEEoverridecommandlockouts
\usepackage{cite}
\usepackage{amsmath,amssymb,amsfonts}
\usepackage{algorithmic}
\usepackage{graphicx}
\usepackage{textcomp}
\usepackage{xcolor}
\usepackage{graphicx}
\usepackage{multirow}
\usepackage{pbox}
\usepackage{flushend}
\usepackage[caption=false]{subfig}
\def\BibTeX{{\rm B\kern-.05em{\sc i\kern-.025em b}\kern-.08em
    T\kern-.1667em\lower.7ex\hbox{E}\kern-.125emX}}
\begin{document}

\newcommand{\multilinebox}[1]{\pbox{\linewidth}{\vspace{.5\baselineskip}#1\vspace{.5\baselineskip}}}
\newcommand\mnote[1]{\textcolor{red}{#1}}

\title{MultiCast: Zero-Shot Multivariate Time Series Forecasting Using LLMs}

\author{\IEEEauthorblockN{Georgios Chatzigeorgakidis}
\IEEEauthorblockA{\textit{``Athena'' Research Center}\\
Greece \\
gchatzi@athenarc.gr}
\and
\IEEEauthorblockN{Konstantinos Lentzos}
\IEEEauthorblockA{\textit{``Athena'' Research Center}\\
Greece \\
klentzos@athenarc.gr}
\and
\IEEEauthorblockN{Dimitrios Skoutas}
\IEEEauthorblockA{\textit{``Athena'' Research Center}\\
Greece \\
dskoutas@athenarc.gr}
}

\maketitle

\begin{abstract}

Predicting future values in multivariate time series is vital across various domains. This work explores the use of large language models (LLMs) for this task. However, LLMs typically handle one-dimensional data. We introduce MultiCast, a zero-shot LLM-based approach for multivariate time series forecasting. It allows LLMs to receive multivariate time series as input, through three novel token multiplexing solutions that effectively reduce dimensionality while preserving key repetitive patterns. Additionally, a quantization scheme helps LLMs to better learn these patterns, while significantly reducing token use for practical applications. We showcase the performance of our approach in terms of RMSE and execution time against state-of-the-art approaches on three real-world datasets.

\end{abstract}


\section{Introduction}
\label{sec:intro}

A time series is a sequence of data points, typically recorded at successive equally spaced intervals of time. These data points can represent various measurements, observations, or readings taken over time, such as temperature readings, stock prices, sales figures, or sensor readings. Time series analysis involves studying the patterns, trends, and relationships present in the data to understand its behavior over time~\cite{shumway2000time}. Time series forecasting predicts future values of a time series based on its past observations. 




Traditional time series forecasting methods have demonstrated considerable efficacy over the years and continue to maintain relevance and widespread adoption in contemporary practice \cite{Zeng_Chen_Zhang_Xu_2023}. In general, these methods can be categorized into linear \cite{9461796}, \cite{DEGOOIJER2006443} and non-linear models \cite{10.1007/11579427_51}, \cite{doi:10.1080/0740817X.2014.999180}.

Arguably, the most popular traditional time series method is AutoRegressive Integrated Moving Average (ARIMA)~\cite{box2015time}. ARIMA consists of three independent components; (i) the AutoRegressive (AR) component assumes that the current value of a time series is a linear combination of its past values, with the addition of a white-noise term; (ii) the Moving Average (MA) component assumes that the current value of a time series variable is a linear combination of past white-noise terms, with no dependence on past values of the variable itself; (iii) the integrated (I) component incorporates differencing to make the time series stationary, allowing for the modeling of nonstationary time series data.

Machine learning and, in particular, deep learning has emerged as a transformative approach in the field of time series forecasting, offering new advances~\cite{10.1145/3533382, doi:10.1089/big.2020.0159, Mahmoud2021, DU2020269}. Moreover, pre-training has been used in deep learning,
to significantly accelerate the training process and increase performance \cite{jiang2022transferability}. In domains such as computer vision and Natural Language Processing (NLP), pre-training facilitates scaling of performance with the availability of data. However, in the context of time series modeling, access to sizable pretraining datasets is often limited.

Large Language Models (LLMs) have emerged as a popular tool for Natural Language Processing (NLP) tasks, and have received considerable attention in recent years. LLMs are pretrained models, trained on vast amounts of text data. Their ability to learn rich representations of language has drawn the attention of the scientific community over the past few years. Specifically, LLMs are quite capable of capturing syntactic, semantic, and contextual information~\cite{zhao2023survey}. Another interesting aspect of LLMs are \textit{emergent abilities}~\cite{wei2022emergent}, which are capabilities that are not explicitly programmed or designed, but rather spontaneously emerge from the complex internal processes of the models. In the past few years, scientists have focused on leveraging the LLMs' potential to solve problems from other fields than NLP. In particular, in time series forecasting, by taking advantage of pre-learned representations of language, LLMs can potentially capture temporal relationships and time series dynamics~\cite{gruver2023large}. However, most works have focused on univariate time series forecasting, requiring either fine-tuning~\cite{chang2023llm4ts}, or a few-shot prompting approach~\cite{xue2023promptcast} (i.e., providing a few examples via prompting to guide the model's behavior for a specific task).

In this work, we examine the utility of LLMs for multivariate time series forecasting via \textit{zero-shot} prompting (i.e., no additional examples are provided). To the best of our knowledge, ours is the first work that addresses this problem.

Our contributions are summarized as follows:
\begin{itemize}
    \item We introduce three dimensional multiplexing techniques to combine all dimensions into a single string, passed to an LLM as input.
    \item We employ SAX quantization on the time series to facilitate inference by the model and to significantly reduce the computational cost and token usage.
    \item We present an experimental evaluation against existing traditional, machine learning, and LLM-based methods for time series forecasting.
\end{itemize}
\section{Related Work}
\label{sec:related}


 LLMs have been applied into many different domains and contexts such as healthcare \cite{schmiedmayer2024llm}, \cite{nashwan2023harnessing}, \cite{wornow2024zero}, 
financial modeling \cite{li2023large}, \cite{yang2023fingpt}, \cite{zhao2024revolutionizing},
and education and research \cite{hosseini2023exploratory}, \cite{moore2023empowering},
as well as in time series data  \cite{jiang2024empowering} for many different tasks and
application domains \cite{zhang2024large}.

The authors of TIME-LLM \cite{jin2024timellm}, introduce a reprogramming framework aimed at adapting
LLMs for time series forecasting without altering their pre-trained structure. TIME-LLM reprograms input time series
into text prototype representations that suit LLMs' capabilities. By introducing Prompt-as-Prefix (PaP), which enriches the input context with natural language instructions, the reprogrammed input is then processed by the frozen LLM. The output is projected to generate time series forecasts updating only lightweight input transformation and output projection parameters, while the backbone language model remains frozen. Scenarios for both short- and long-term are addressed, as well as few- and one-shot learning. 

LLMTIME~\cite{gruver2023large} is the first approach to apply zero-shot forecasting on time series using LLMs. The authors argue that the output of LLMs when predicting digit-by-digit follows a multimodal distribution, which fits well in the case of time series. To apply forecasting, the time series values are tokenized and rescaled to a \textit{predefined} number of digits to use fewer tokens. Then, to apply forecasting, the time series with their tokenized values separated by commas are passed to the model. Notably, the model's output is limited to producing only digits and commas (i.e., $[0-9,]$). At each time step, a predefined number of samples is drawn and the final forecast is built using the median of all samples after descaling the outputted values. 

Despite the potential of LLMs for time series forecasting, there are several limitations that need to be addressed.

\begin{itemize}
\item \textit{No multivariate support}: Most current approaches using LLMs for time series forecasting focus on univariate time series data. This limitation restricts the applicability of LLMs to certain types of time series data. 

\item \textit{Fine-tuning requirement}: Fine-tuning can be time-consuming and computationally expensive, particularly for large models. It also requires a substantial amount of training data, which may not always be available.

\item \textit{Number of tokens required}: LLMs are extremely large models, capable of efficiently running on computers equipped with GPUs of high capacity in RAM. Thus, their broad availability depends on services that host such models, which usually charge queries by token. Consequently, very large queries (e.g., a large time series in our context) would be rather expensive to run.


\end{itemize}

\section{MultiCast}
\label{sec:methods}
In the following, we describe our approach to zero-shot multivariate time series forecasting using LLMs. First, we go through the three separate token multiplexing approaches that we propose. Then, we describe our approach to reducing complexity using the SAX representation.

\subsection{Dimensional Multiplexing}
\label{subsec:dim_mult}
\begin{figure}[htbp]
\centerline{\includegraphics[width=0.45\textwidth]{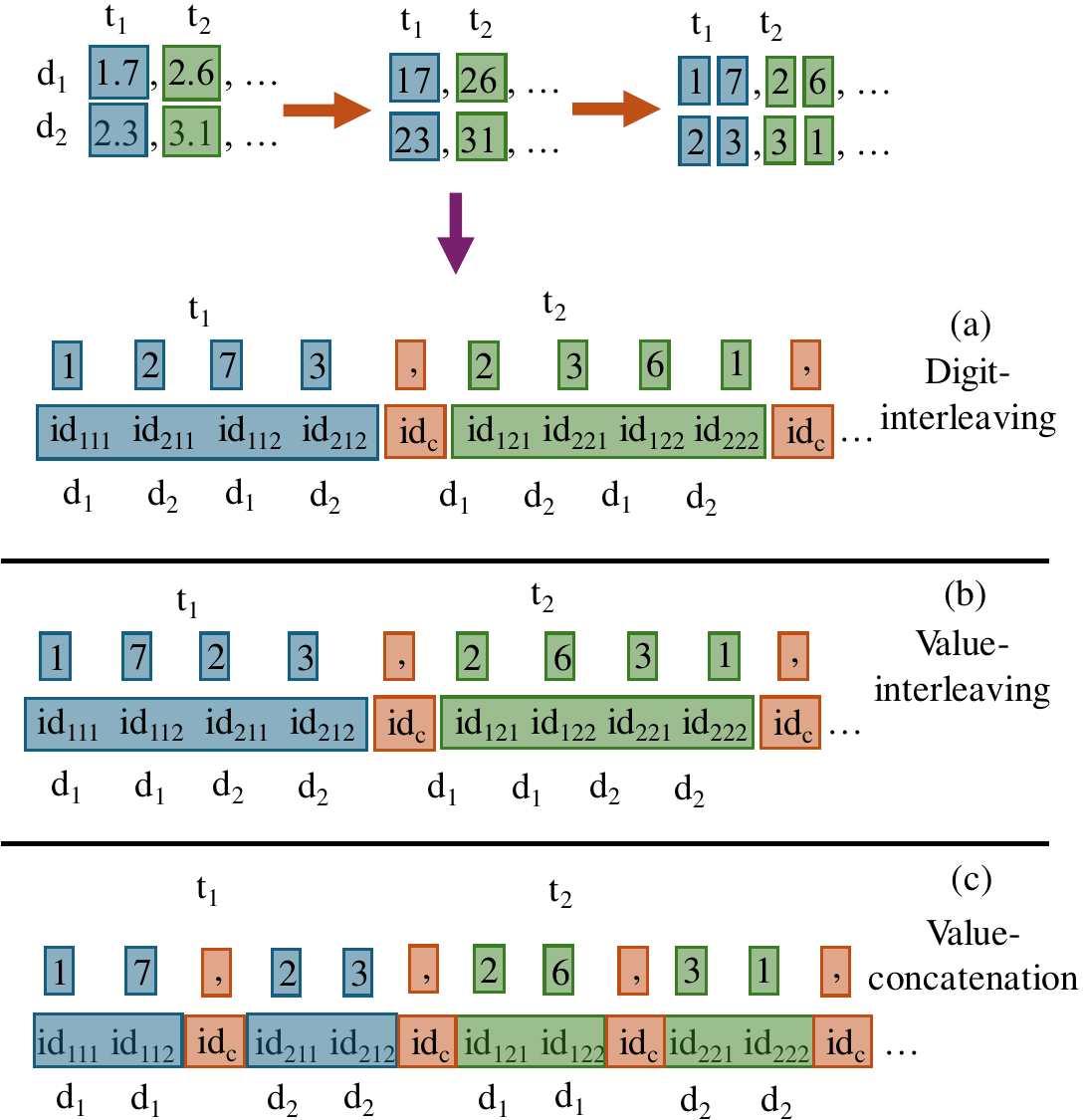}}
\caption{The three token multiplexing techniques.}
\label{fig:multiplex}
\end{figure}

The dimensional multiplexing process takes place after each dimension has been rescaled to avoid decimals. Then, each digit is treated separately. An example of this process is illustrated in the top row of Figure~\ref{fig:multiplex}. Of course, depending on the LLM used, its tokenizer must be adapted accordingly, as discussed in~\cite{gruver2023large}. After multiplexing, the tokens are replaced with their corresponding corpus id before being passed onto the model for inference. When the model produces the output, this process is reversed to obtain the final result. We introduce three separate dimensional multiplexing techniques, namely (i) \textit{digit-interleaving}, (ii) \textit{value-interleaving}, and (iii) \textit{value-concatenation}.

\subsubsection{Digit-Interleaving}
After each dimension has been rescaled, the \textit{Digit-Interleaving} (DI) multiplexing technique places the digits of each dimension per timestamp interchangeably. This is exemplified in Figure~\ref{fig:multiplex}a. Consider a 2-dimensional time series. Specifically, $d_1 = [1.7, 2.6, ...]$ and $d_2 = [2.3, 3.1, ...]$ are the two dimensions (i.e., we only show the first two timestamps for brevity). After rescaling, the dimensions become $d_1 = [17, 26, ...]$ and $d_2 = [23, 31, ...]$, respectively. Then, as described previously, each digit is considered a separate token. Before being assigned the corresponding corpus id, tokens are interchangeably placed per dimension for each timestamp, reducing the dimensions to 1. The resulting series in the example would be $d = [1273, 2361, ...]$. Then, each digit (token) and comma are assigned with the corresponding id. This technique attempts to take advantage of the fact that, in many multivariate time series, all values are correlated and similarly scaled. Such an example are z-normalized series, which have zero mean with values differing a few standard deviations from it. In such a case, the left-wise digits per dimension will be all placed first; since the model is producing the output token-by-token, this can help it infer the correct scaling of the series. More formally, DI multiplexing can be formulated as follows.


\footnotesize
\begin{equation}
\begin{split}
    I_d = \{t_{111} ... t_{d11} \quad  t_{11b} ... t_{d1b}\} \quad t_c \quad \{t_{1n1} ... t_{dn1} \quad t_{1nb} ... t_{dnb}\}
\end{split}
\end{equation}
\normalsize

where $d$ is the number of dimensions, $b$ the predefined number of digits per timestamp, and $n$ the time series length.

\subsubsection{Value-Interleaving}
Figure~\ref{fig:multiplex}b shows the \textit{Value-Interleaving} (VI) dimensional multiplexing technique. This time, instead of interchangeably placing the digits per timestamp and dimension, we place the whole values of each dimension per timestamp one after the other. Thus, in the example, the 1-dimensional result will be $d = [1723, 2631, ...]$. Intuitively, this technique is more suitable in cases where the dimensions of the series are on a different scale. We expect the model to be able to distinguish between the different dimensions --especially when they differ in scale--, and manage to internally demultiplex the input before inference. The VI multiplexing can be formulated as follows.

\footnotesize
\begin{equation}
\begin{split}
    I_ts = \{t_{111} ... t_{11b} \quad  t_{d11} ... t_{d1b}\} \quad t_c \quad \{t_{1n1} ... t_{1nb} \quad t_{dn1} ... t_{dnb}\}
\end{split}
\end{equation}
\normalsize

where $d$ is the number of dimensions, $b$ the predefined number of digits per timestamp, and $n$ the time series length.

\subsubsection{Value-Concatenation}
Finally, Figure~\ref{fig:multiplex}b shows the \textit{Value-Concatenation} (VC) dimensional multiplexing technique which is an extension of the value-interleaving technique; for each timestamp, we now place the values of each dimension separated by commas, thus considering them as different values (e.g., in the figure, the 1-dimensional result will be $d = [17, 23, 26, 31, ...]$. We expect this to further faciliate the internal demultiplexing by the model before detecting any patterns. The VC multiplexing can be formulated as follows.

\footnotesize
\begin{equation}
\begin{split}
    I_td = \{t_{111} ... t_{11b}\} \hspace{0.1em} t_c \hspace{0.1em} \{t_{d11} ... t_{d1b}\} \hspace{0.1em} t_c \hspace{0.1em} \{t_{1n1} ... t_{1nb}\} \hspace{0.1em} t_c \hspace{0.1em} \{t_{dn1} ... t_{dnb}\}
\end{split}
\end{equation}
\normalsize

where $d$ is the number of dimensions, $b$ the predefined number of digits per timestamp and $n$ the length of the time series. Of course, in all cases, upon receiving the multiplexed output from the model, the tokens must be properly decoded, demultiplexed, and brought back to their initial scale for each dimension, depending on the selected technique. A significant advantage of this multiplexing technique against forecasting each dimension separately is the fact that multivariate time series tend to have high interdimensional correlations (e.g., temperature and humidity in weather data). We expect that providing them altogether in the model can lead to the detection of such interdimensional patterns, yielding better results.

\subsection{Quantization Using SAX}
\label{subsec:quant_sax}
The {\em Symbolic Aggregate approXimation (SAX)} is a multi-resolution representation of a time series introduced in \cite{shieh2008kdd}. It can be derived from its {\em Piecewise Aggregate Approximation} (PAA) \cite{keogh2001paa,faloutsos2000vldb} by quantizing the PAA segments on the $v$-axis. A time series is first transformed into a PAA representation of $w$ segments with real-valued coefficients. To obtain a $SAX$ word for a time series, these coefficients are discretized along the value axis using {\em breakpoints} assuming a $\mathcal{N}(0,1)$ Gaussian distribution that enables the generation of equiprobable symbols for a given cardinality. Although bitwise representations were used for these symbols in the original paper, other encoding types are also possible. Two such popular alternatives are using alphabetical characters or digits for each symbol.



Forecasting time series is an inherently difficult task due to the nature of the data. This is also the case for zero-shot foreasting using LLMs, since (i.e., as also described in~\cite{gruver2023large}) they have to infer a sequence of tokens for each timestamp, thus simulating a multi-modal distribution. This becomes even harder when applying the above-mentioned dimensional multiplexing techniques. Also, for large time series, such a process becomes significantly more computationally intensive; plus, it requires many tokens, which, depending on the application, can be rather expensive to infer according to currently LLM pricing policies. To alleviate these issues, we quantize the time series across all dimensions in both axes using the SAX representation, before applying tokenization. We support two different quantization types, either using an alphabetical or a digital SAX alphabet. Now, each value per timestamp is consisted of only one token instead of multiple. For example, the time series in Figure~\ref{fig:multiplex} could become $d_1^{sax} = [a,b,...]$ and $d_2^{sax} = [b,c,...]$ after alphabetical quantization. We expect that it will be easier for the model to detect patterns when dealing only with one token per timestamp. 
\section{Experimental Evaluation}
\label{sec:exp}

This section presents the results of our experiments. We first explain how we set up our tests and assess the suggested methods.

\subsection{Experimental Setup}
\label{subsec:exp_setup}

\subsubsection{System}
We used Python and the Hugging Face API\footnote{https://huggingface.co/}. The experiments were run on a server with an AMD Ryzen Threadripper 3960X 24-core CPU and 256GB memory. The experiments were run on CPU. 


\subsubsection{Datasets}
We employ three real-world multivariate time series datasets. 


\textit{Gas Rate}: This is a 2-dimensional dataset containing carbon dioxide (CO$_2$) emissions. The first dimension contains the input CO$_2$ measurements (ft3/min) in a gas furnace. The second dimension contains the output CO$_2$ percentage. The dataset is obtained from the darts library\footnote{https://unit8co.github.io/darts}. Of course, the two dimensions are correlated, which makes this dataset ideal for multivariate forecasting.

\textit{Electricity}: This multivariate time series is part of the Electricity Transformer Dataset (ETDataset)\footnote{https://github.com/zhouhaoyi/ETDataset}. It contains hourly measurements of various metrics, which were resampled on a 3-day basis, for a total of 242 timestamps. From this dataset, we extracted 3 dimensions of electricity measurements, specifically the High UseFul Load (HUFL), High UseLess Load (HULL), and Oil Temperature (OT). Again, the dimensions are correlated; specifically, OT is used as a target variable in regression problems.

\textit{Weather}: The weather dataset was generated by the Max Planck Institute\footnote{https://www.bgc-jena.mpg.de/wetter/} and contains 21 weather-related metrics obtained from a weather station located in Germany. From the 21 variables, we extracted the air temperatures (Tlog) measured in Celsius degrees, the water vapor concentration (H2OC) measured in mmol/mol, the saturation water vapor pressure (VPmax), measured in mbar, and the potential temperature (Tpot) measured in Kelvin degrees. Again, being weather-related, all dimensions are correlated.

\begin{table}[ht!]
\caption{Datasets.}
\vspace{-12pt}
\begin{center}
\scriptsize
\renewcommand{\arraystretch}{1.2}
\begin{tabular}{|c|c|c|}
\hline
\textbf{Dataset} & \textbf{Dimensions} & \textbf{Length} \\
\hline
Gas Rate & 2 & 296 \\
Electricity & 3 & 242 \\ 
Weather & 4 & 217 \\
\hline
\end{tabular}
\label{tab:datasets}
\end{center}
\vspace{-15pt}
\end{table}

\begin{table}[ht!]
\caption{Parameters.}
\vspace{-12pt}
\begin{center}
\scriptsize
\renewcommand{\arraystretch}{1.2}
\begin{tabular}{|c|c|}
\hline
\textbf{Parameter} & \textbf{Range} \\
\hline
Dimensions & \textbf{2}, 3, 4 \\
Number of samples & \textbf{5}, 10, 20  \\
SAX segment length & 3, \textbf{6}, 9  \\
SAX alphabet size & \textbf{5}, 10, 20 \\
\hline
\end{tabular}
\label{tab:params}
\end{center}
\end{table}

\subsubsection{Competitors}
We evaluate the following methods: 

\begin{itemize}
    \item \textbf{MultiCast (DI)}: MultiCast using the digit-interleaving dimensional multiplexing method.
    \item \textbf{MultiCast (VI)}: MultiCast using the value-interleaving dimensional multiplexing method on the same value.
    \item \textbf{MultiCast (VC)}: MultiCast using the value-concatenation dimensional multiplexing method on consecutive values.
    \item \textbf{LLMTIME}: The state-of-the-art in LLM-based zero-shot time series forecasting (i.e., applied in each dimension separately).
    \item \textbf{ARIMA}: Autoregressive Integrated Moving Average (ARIMA) is one of the most widely used univariate time series forecasting methods.
    \item \textbf{LSTM}~\cite{lstm97}: Termed as Long-Short-Term Memory (LSTM), LSTMs are Recurrent Neural Networks (RNNs) designed to handle the vanishing gradient problem. This ability allows LSTMs to learn and remember information over time, making them ideal for time series forecasting. LSTMs have been used successfully for multivariate time series forecasting~\cite{alhirmizy2019multivariate, ju2021multivariate}.
\end{itemize}

 
\subsubsection{Parameters}
The parameters utilized in our experimental assessment are listed in Table~\ref{tab:params}. For each parameter, we performed tuning tests to establish their ranges and default values, which are highlighted in bold within the table. More specifically, the \textit{dimensions} parameter corresponds to the dimensionality of each dataset; the \textit{number of samples} only applies to the LLM-related models and is the number of inference values taken for each timestamp; the \textit{SAX segment length} is the level of quantization on the x-axis, which determines the level of compression of a time series; the \textit{SAX alphabet size} is the level of quantization on the y-axis, as performed by the SAX method. Regarding the LSTM parametrization, we performed a grid search, which yielded a 1-hidden-level network of 128 units and a dropout rate of 0.2. It was trained for 30 epochs using the Adam~\cite{kingma2014adam} optimizer with the Mean Squared Error (MAE) as loss function.

\subsubsection{Metrics}
In accordance with standard practices in time series forecasting, the Root Mean Squared Error (RMSE) metric was employed to evaluate our methods. RMSE is formulated as $\sqrt{\Sigma_{i=1}^{n}{({y_i-\hat{y}_i})^2}/{n}}$, where $y_i$ is the actual value, $\hat{y}_i$ is the predicted value at timestamp $i$ and $n$ is the number of timestamps on which forecasting was applied.

\subsection{LLM Model Selection}
MultiCast can be used with any LLM to apply multivariate time series forecasting. In the following, we evaluate its accuracy using LLaMA2 (i.e., the 7B parameter variant) and Phi-2~\cite{javaheripi2023phi} as back-end models. LLaMA2 is one of the most popular LLMs that achieves good performance with fewer parameters. Phi-2 is a math-oriented LLM (i.e., 2.7B parameters), tailored to solving math problems. Table~\ref{tab:models} lists the forecast RMSE for the Gas Rate data set in both dimensions for both LLMs. In both cases, the VI variant of MultiCast was used. LLaMA2 achieves better performance (i.e., approx. twice as good) in all cases. This can be attributed to the fewer parameters of the Phi-2 model; while it is math-oriented and quite capable of solving complex problems described by textual prompting, it seems to not properly detect the patterns in the series, leading to larger errors.

\begin{table}[ht!]
\caption{LLM model comparison.}
\vspace{-12pt}
\begin{center}
\scriptsize
\renewcommand{\arraystretch}{1.2}
\begin{tabular}{|c|c|c|}
\hline
\multirow{2}{1cm}{\textbf{Model}} & \multicolumn{2}{c|}{\textbf{Dimension}} \\
\cline{2-3}
& \textbf{GasRate} & \textbf{CO2} \\
\hline
MultiCast (LLaMA2 / 7B) & \textbf{1.154} & \textbf{2.71} \\
MultiCast (Phi-2 / 2.7B) & 2.106 & 4.676  \\
\hline
\end{tabular}
\label{tab:models}
\end{center}
\end{table}

\begin{figure}[ht!]
    \centering
    \setcounter{subfigure}{0}
    \subfloat[LLaMA2 GasRate prediction]{\includegraphics[width=0.4\textwidth]{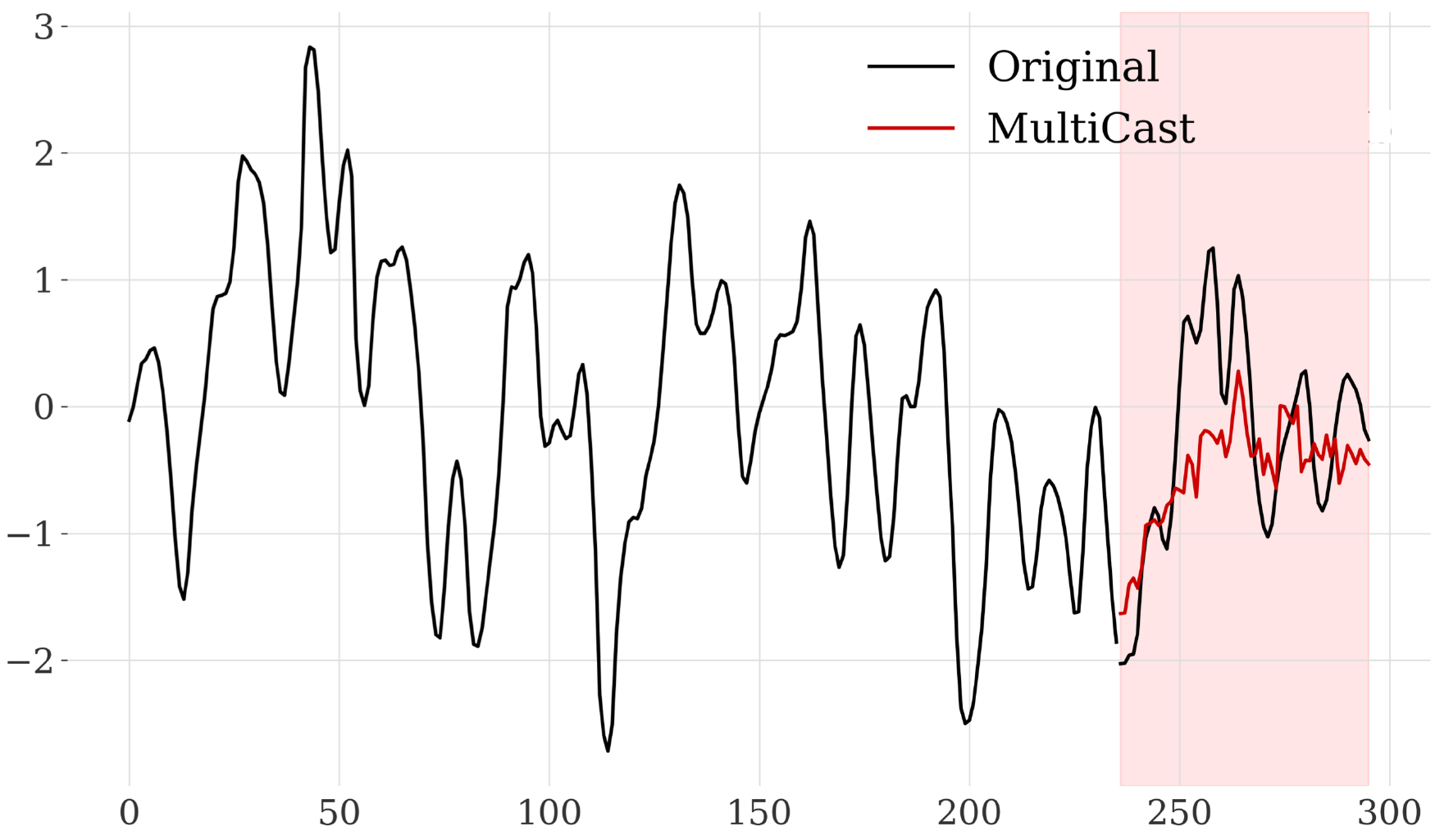}\label{subfig:llama}} \\
    \vspace{-10pt}
    \subfloat[Phi-2 GasRate prediction]{\includegraphics[width=0.4\textwidth]{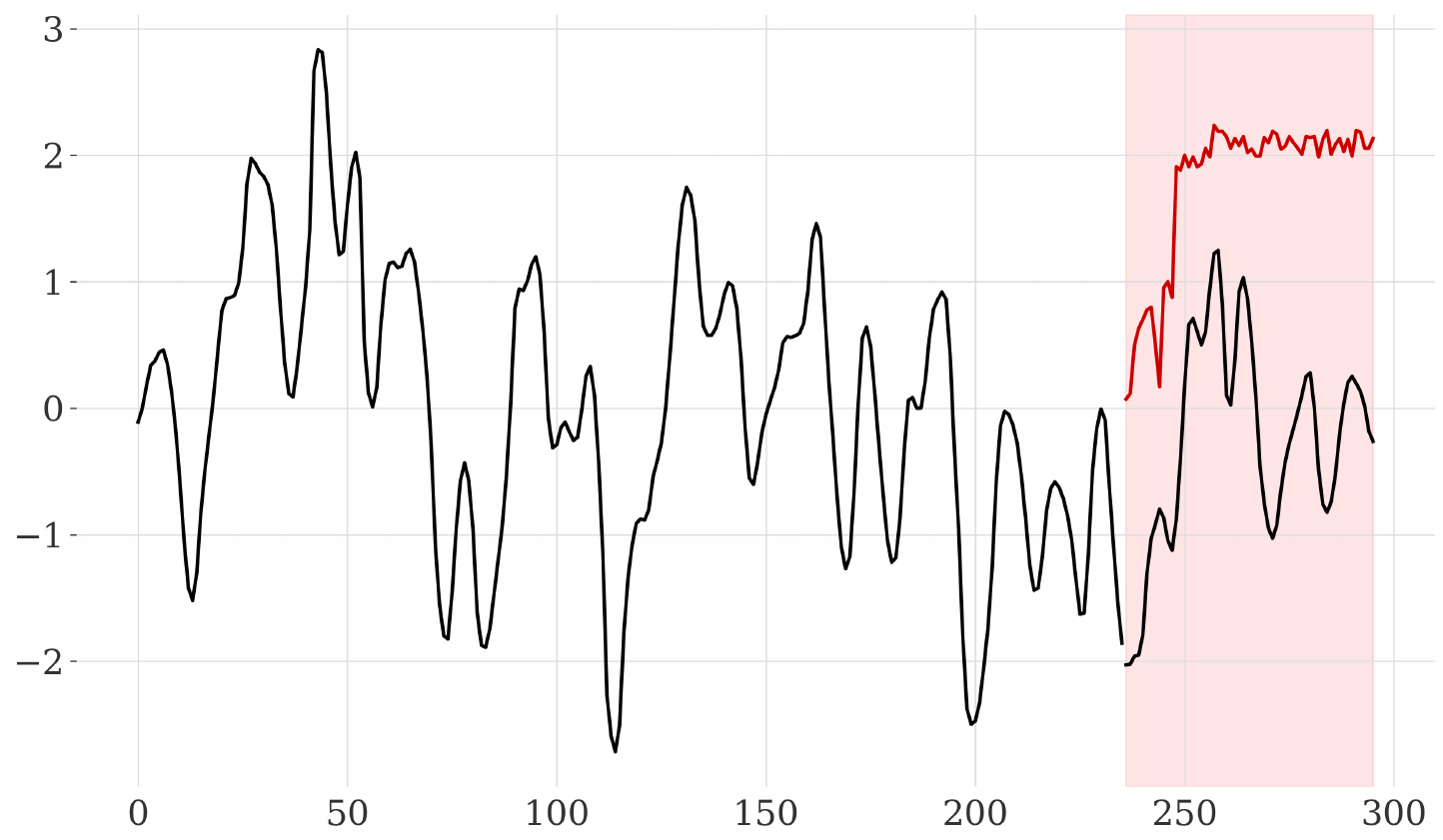}\label{subfig:phi2}} \\
    \caption{Comparison of the two models.}
    \label{fig:supported_models}
    \vspace{-6pt}
\end{figure}

Figures~\ref{fig:supported_models}a and b depict two indicative examples of forecasting the first dimension of the Gas Rate dataset using the LLaMA2 and Phi-2 models, respectively. Clearly, the LLaMA2 model performs better, being able to properly follow the upward trend of the time series and even infer two local maxima of the original time series. Phi-2, on the other hand, fails to accurately forecast the time series in this dimension; while it seems to successfully detect the upward trend, its entire output is shifted 1 to 2 units on the y-axis. Since LLaMA2 seems to perform significantly better in all cases, for the rest of the experiential evaluation, we will be using LLaMA2 as the back-end model for MultiCast.

\subsection{Forecasting Accuracy}
\label{subsec:forecast_acc}
Next, we compare the prediction precision in terms of RMSE of all MultiCast variants against the rest of the competitor approaches. Table~\ref{tab:forecasting_gas} lists the results for the Gas Rate dataset. To better comprehend the insights behind the results and acquire knowledge regarding the differences in forecasting ability between the LLM-based models and the rest of the competition, for each dimension, we denote the first best overall performance using bold font and the second best using italic font. Interestingly, for the GasRate dimension, the best overall approach was LLMTIME ($0.703$), followed by MultiCast (DI) with $0.781$. The LLM-based approaches all seem to cope well with detecting the underlying patterns for this dimension, thus producing good results. The case is different for the second dimension (CO2), where the conventional methods seem to yield a better overall performance, with ARIMA being the best (2.63). MultiCast (VI) was the second-best overall and the best LLM-based performer (2.71).

\begin{table}[ht!]
\caption{Forecasting RMSE for the Gas Rate dataset.}
\vspace{-12pt}
\begin{center}
\scriptsize
\renewcommand{\arraystretch}{1.2}
\begin{tabular}{|c|c|c|}
\hline
\multirow{2}{0.8cm}{\textbf{Model}} & \multicolumn{2}{c|}{\textbf{Dimension}} \\
\cline{2-3}
& \textbf{GasRate} & \textbf{CO2} \\
\hline
MultiCast (DI) & \textit{0.781} & 4.639 \\
MultiCast (VI) & 1.154 & \textit{2.71}  \\
MultiCast (VC) & 0.965 & 3.626  \\
LLMTIME & \textbf{0.703} & 2.75  \\
ARIMA & 0.92 & \textbf{2.63}  \\
LSTM & 1.122 & 3.89  \\
\hline
\end{tabular}
\label{tab:forecasting_gas}
\end{center}
\end{table}

\begin{table}[ht!]
\caption{Forecasting RMSE for the Electricity dataset.}
\vspace{-12pt}
\begin{center}
\scriptsize
\renewcommand{\arraystretch}{1.2}
\begin{tabular}{|c|c|c|c|}
\hline
\multirow{2}{0.8cm}{\textbf{Model}} & \multicolumn{3}{c|}{\textbf{Dimension}} \\
\cline{2-4}
& \textbf{HUFL} & \textbf{HULL} & \textbf{OT} \\
\hline
MultiCast (DI) & 5.914 & 1.444 & 9.198 \\
MultiCast (VI) & 8.63 & 1.882 & 13.752 \\
MultiCast (VC) & \textbf{2.424} & 1.913 & 10.230   \\
LLMTIME & \textit{4.299} & \textit{1.432} & \textit{7.543}   \\
ARIMA & 7.063 & 1.572 & \textbf{4.181}   \\
LSTM & 4.892 & \textbf{1.43} & 8.740   \\
\hline
\end{tabular}
\label{tab:forecasting_elect}
\end{center}
\end{table}

\begin{table}[ht!]
\caption{Forecasting RMSE for the Weather dataset.}
\vspace{-12pt}
\begin{center}
\scriptsize
\renewcommand{\arraystretch}{1.2}
\begin{tabular}{|c|c|c|c|c|}
\hline
\multirow{2}{0.8cm}{\textbf{Model}} & \multicolumn{4}{c|}{\textbf{Dimension}} \\
\cline{2-5}
& \textbf{Tlog} & \textbf{H2OC} & \textbf{VPmax} & \textbf{Tpot} \\
\hline
MultiCast (DI) & 3.711 & 2.43 & 3.025 & 6.888 \\
MultiCast (VI) & \textit{3.26} & 2.122 & \textbf{2.387} & 11.352 \\
MultiCast (VC) & 4.983 & 3.819 & 5.776 & \textit{5.993} \\
LLMTIME & \textbf{3.14} & \textbf{1.746} & 4.044 & 6.981 \\
ARIMA & 3.324 & 2.686 & 4.331 & 6.067 \\
LSTM & 3.524 & \textit{1.796} & \textit{2.708} & \textbf{5.559} \\
\hline
\end{tabular}
\label{tab:forecasting_weather}
\end{center}
\end{table}

\begin{figure}[ht!]
    \centering
    \setcounter{subfigure}{0}
    \subfloat[Multicast for the gas rate dataset (GasRate).]{\includegraphics[width=0.4\textwidth]{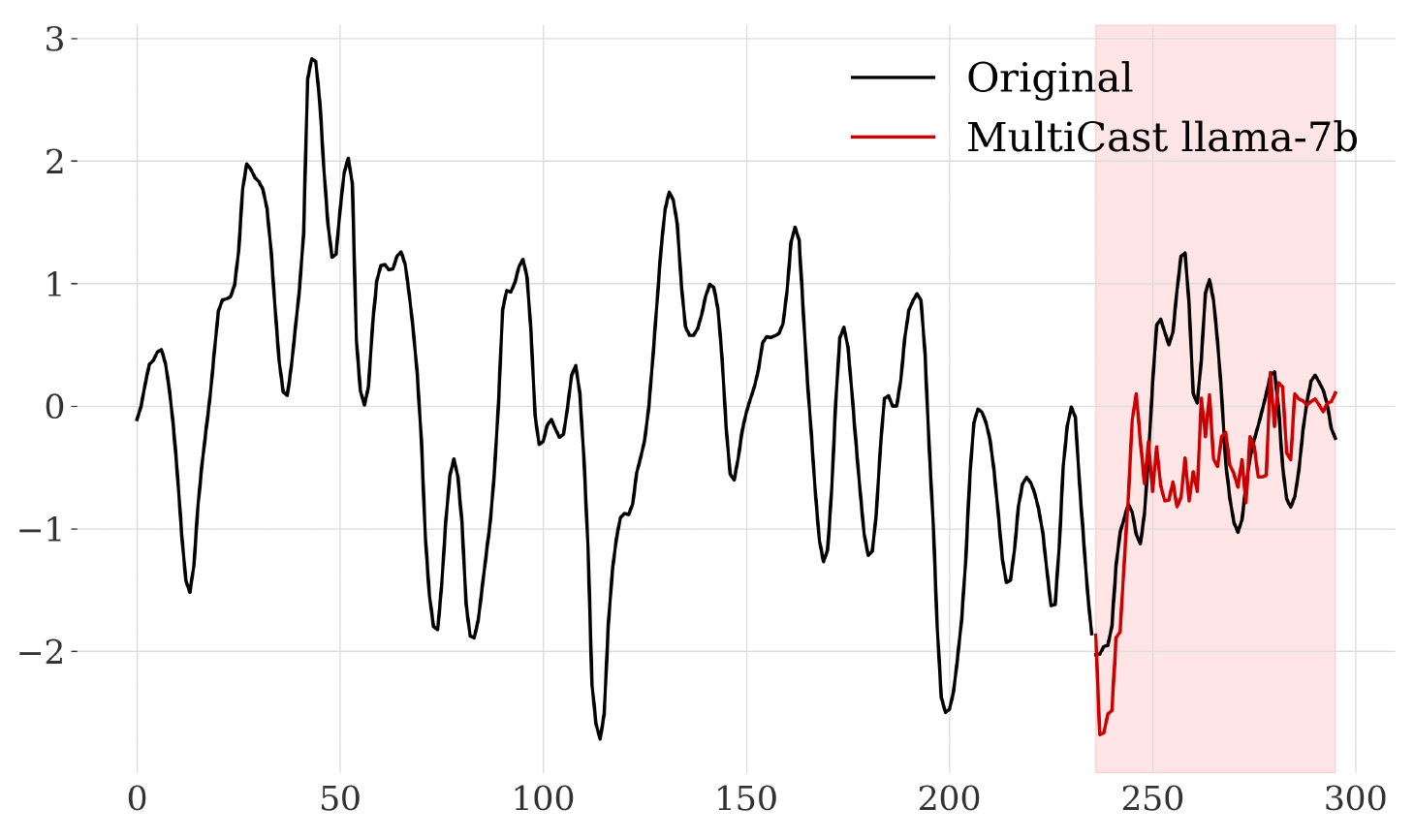}\label{subfig:multi_gas}} \\
    \vspace{-10pt}
    \subfloat[ARIMA for the gas rate dataset (GasRate).]{\includegraphics[width=0.4\textwidth]{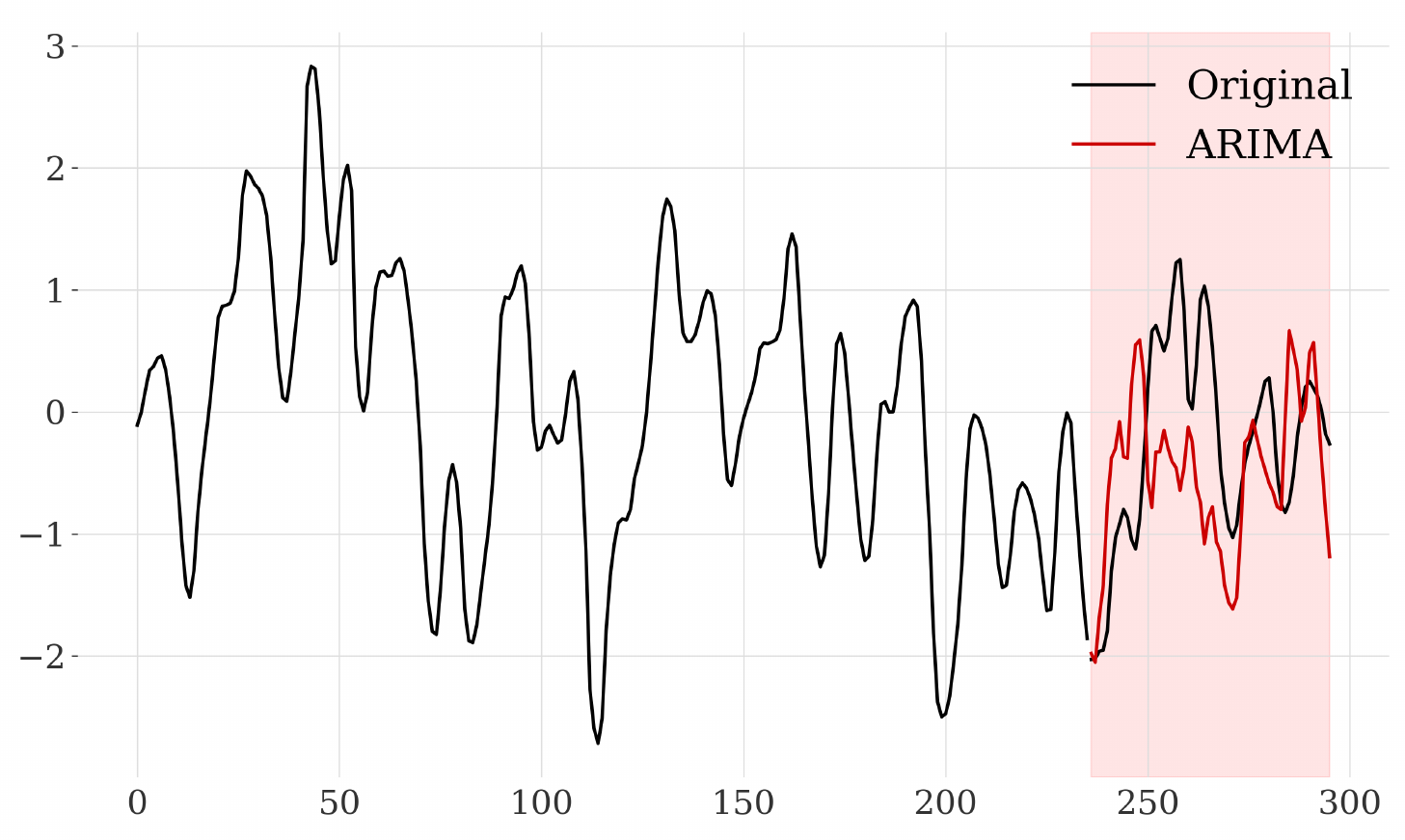}\label{subfig:arima_gas}} \\
    \caption{MultiCast (DI) versus ARIMA for the GasRate dimension.}
    \label{fig:comp_gas_rate}
    \vspace{-6pt}
\end{figure}

Figure~\ref{fig:comp_gas_rate} depicts two indicative forecast outputs of the best MultiCast approach (DI) for the first dimension of the Gas Rate data set against the corresponding ARIMA result. Both seem to yield a good result here; MultiCast seems to properly detect a continuously upward trend in the time series; however, the result seems to have larger variance than that of the original time series. On the other hand, the ARIMA approach does not clearly follow the upward trend; however, its variance seems to be on par with the one of the original time series.

Table~\ref{tab:forecasting_elect} lists the results for the Electricity dataset. For the HUFL dimension, MultiCast (VC) seems to yield significantly better RMSE than the rest of the approaches. However, the rest of the MultiCast variants do not cope as well. For the HULL dimension, all approaches seem to produce good results, with LLMTIME achieving the best RMSE. Finally, for the OT dimension, ARIMA performs significantly better than the competition. The MultiCast approaches do not perform well. LLMTIME is the best among LLM-based models. This suggests a possible drop in the performance of MultiCast as the dimensionality of the time series increases since there is the extra step of demultiplexing the input that the LLMs must infer. However, the error in the best LLM-based model (9.198) is very close to that of the LSTM model (8.740).

\begin{figure}[ht!]
    \centering
    \setcounter{subfigure}{0}
    \subfloat[Multicast for the electricity dataset (HUFL).]{\includegraphics[width=0.4\textwidth]{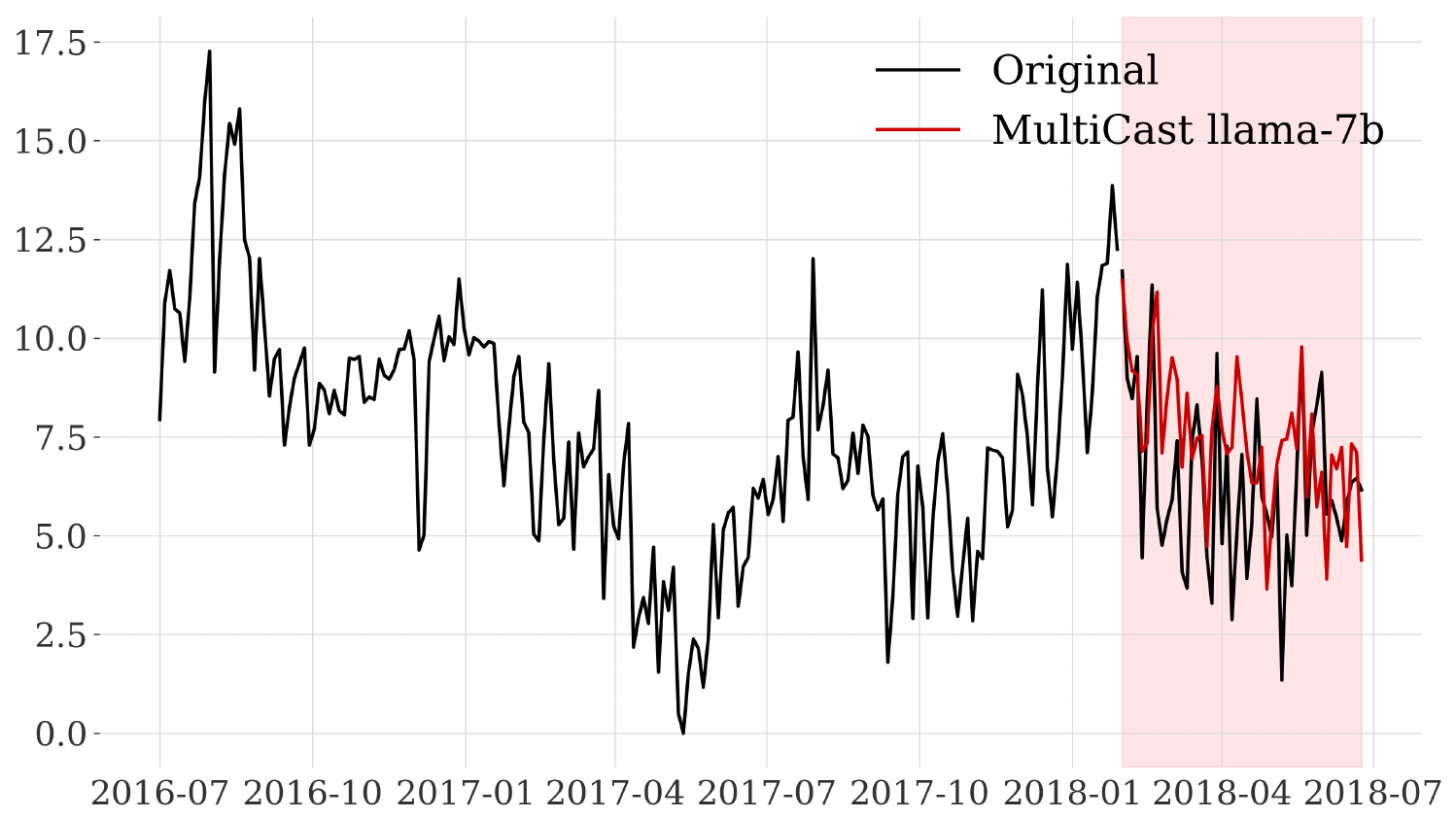}\label{subfig:multi_electr}} \\
    \vspace{-10pt}
    \subfloat[LSTM for the electricity dataset (HUFL).]{\includegraphics[width=0.4\textwidth]{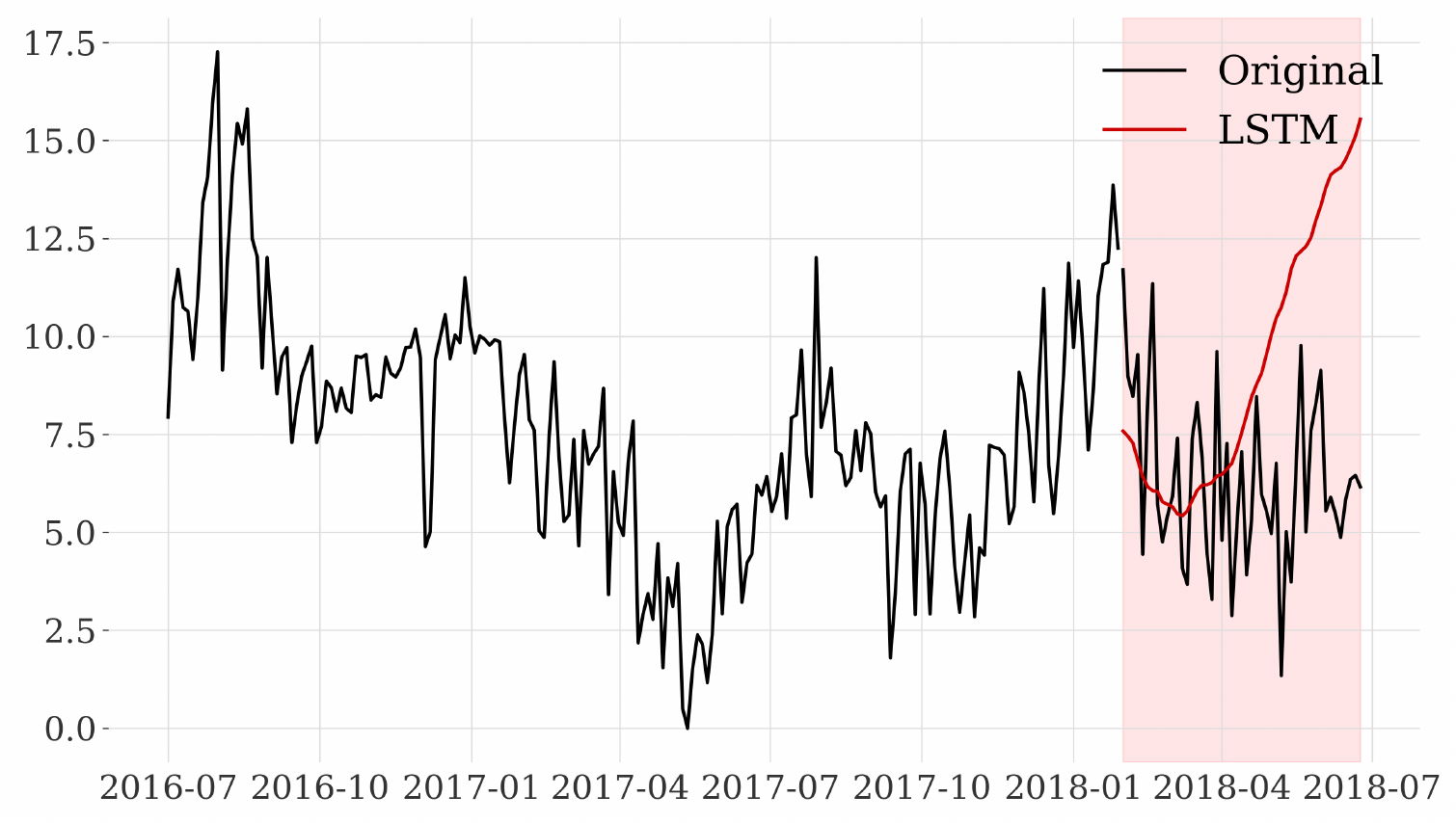}\label{subfig:lstm_electr}} \\
    \caption{MultiCast (VC) versus LSTM for the HUFL dimension.}
    \label{fig:comp_electr}
    \vspace{-6pt}
\end{figure}

Figure~\ref{fig:comp_electr} illustrates an indicative example of the MultiCast (VC) forecast output (Figure~\ref{subfig:multi_electr}) against the LSTM (Figure~\ref{subfig:lstm_electr}) for the HUFL dimension of the electricity data set. Clearly, MultiCast manages to correctly infer both the trend and variance of the time series. On the other hand, the LSTM seems to perform rather poorly, falsely yielding a non-existent linear upward trend.

The RMSE results for the Weather dataset are listed in Table~\ref{tab:forecasting_weather}. LLMTIME achieves the best performance in the Tlog dimension, though, all approaches except MultiCast (VC) are close. This is also the case for the H2OC dimension. For the VPmax dimension, the best overall approach was MultiCast (VI), with MultiCast (VC) again performing worse than the rest. However, this is reversed in the Tpot dimension, where the MultiCast variant (VC) yields the best performance among all LLM-based approaches. LSTM is the better performer in this dimension. Notice that the degradation in forecasting accuracy for more dimensions is not present in this case; the MultiCast variants are all either close to, or outperform the rest in all dimensions. Another key takeaway here is that the optimal multiplexing method differs from dimension to dimension and from dataset to dataset. A comprehensive analysis on which dataset characteristics cause this behavior is an interesting future work.

\begin{figure}[ht!]
    \centering
    \setcounter{subfigure}{0}
    \subfloat[Multicast for the weather dataset (Tlog).]{\includegraphics[width=0.4\textwidth]{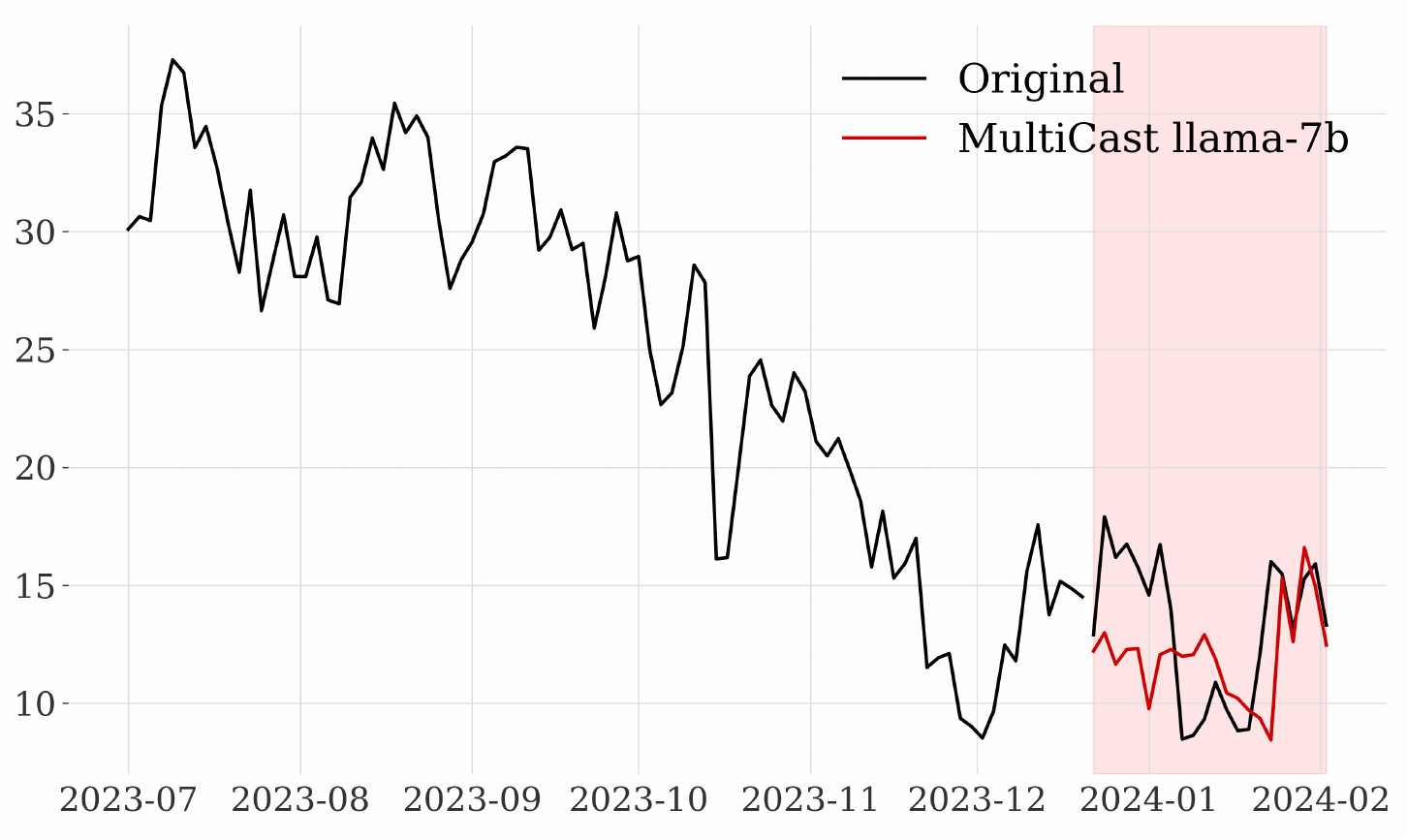}\label{subfig:multi_weather}} \\
    \vspace{-10pt}
    \subfloat[ARIMA for the weather dataset (Tlog).]{\includegraphics[width=0.4\textwidth]{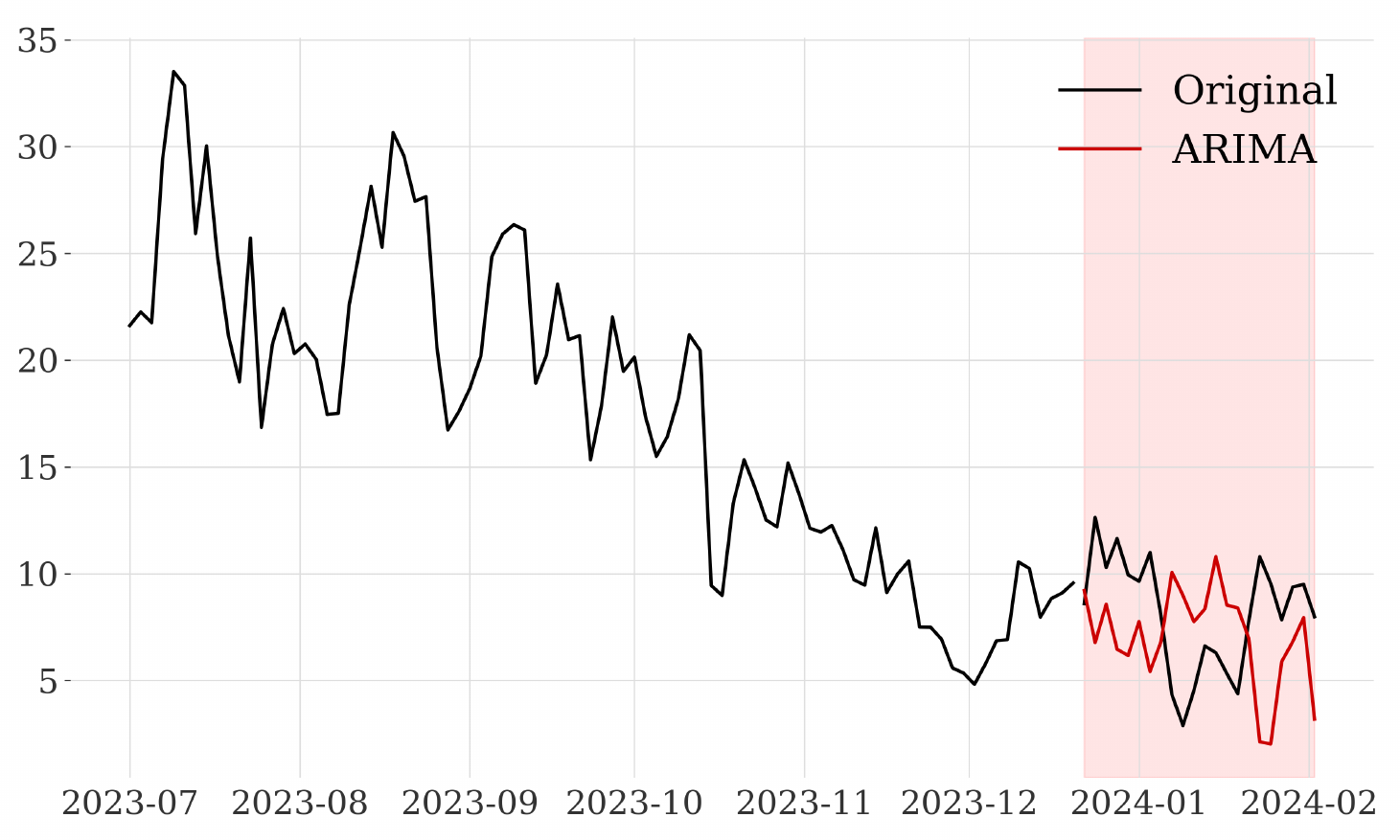}\label{subfig:arima_weather}} \\
    \caption{MultiCast (VI) versus ARIMA for the Tlog dimension.}
    \label{fig:comp_weather}
    \vspace{-6pt}
\end{figure}

As in the rest of the cases, an indicative example of MultiCast against a conventional method is illustrated in Figure~\ref{fig:comp_weather}. Clearly, the DI variant of MultiCast (Figure~\ref{subfig:multi_weather}) yields better results than ARIMA (Figure~\ref{subfig:arima_weather}) here, able to accurately estimate the upward trend and fluctuation at the end of the time series.

Overall, we notice a trade-off when using MultiCast for multivariate time series forecasting, as opposed to LLMTIME. Forecasting each dimension separately using LLMTIME will completely ignore the interdimensional correlations, which is not desirable in such scenarios. On the other hand, MultiCast poses an additional challenge to the LLM models, which now have to also infer the demultiplexing of the dimensions. Both cases hinder the accuracy of the obtained result. Having in mind the interesting aspect of emergent abilities, we argue that using very large LLMs (e.g., GPT-4, Gemini) will further improve MultiCast's performance.

\subsection{Increasing Number of Samples}
Table~\ref{tab:num_samples} lists the accuracy in terms of RMSE of all LLM-based models for an increasing number of samples. As a reminder, all LLM-based models draw several samples of the values of each timestamp, and the final estimated value is derived by computing the median among all samples. The LLMTIME approach seems to produce better results for 5 and 10 samples. Interestingly, the error worsens for 20 samples. This could be because the inherent variance of the produced series tends to be averaged out as we draw more samples. However, this is not the case for the MultiCast method; all three approaches seem to produce better results for more samples, and the MultiCast DI variant achieves the best performance for 20 samples. A drawback of drawing many samples is the performance deterioration in execution time (i.e., each execution time is listed below each corresponding RMSE in Table~\ref{tab:num_samples}). Notice that, in all cases, the execution time doubles when the number of samples is doubled, which is expected since the model must infer twice as many tokens. Interestingly, the LLMTIME requires slightly less total time (i.e., sum of time needed per dimension) than its MultiCast counterparts, since the latter also need to infer the multiplexing/demultiplexing of the tokens.

\begin{table}[ht!]
\caption{Performance for an increasing number of samples.}
\vspace{-12pt}
\begin{center}
\scriptsize
\renewcommand{\arraystretch}{1.2}
\begin{tabular}{|c|c|c|c|}
\hline
\multirow{2}{0.8cm}{\textbf{Method}} & \multicolumn{3}{c|}{\textbf{Number of samples}} \\
\cline{2-4}
& \textbf{5} & \textbf{10} & \textbf{20} \\
\hline
MultiCast (DI) & \multilinebox{0.781 \\ 1036 sec} & \multilinebox{0.762 \\ 2050 sec} & \multilinebox{\textbf{0.592} \\ 4159 sec} \\
MultiCast (VI) & \multilinebox{0.965 \\ 1041 sec} & \multilinebox{1.302 \\ 2068 sec} & \multilinebox{0.877 \\ 4131 sec}  \\
MultiCast (VC) & \multilinebox{1.154 \\ 1168 sec} & \multilinebox{0.704 \\ 2468 sec} & \multilinebox{0.63 \\ 4981 sec}  \\
LLMTIME & \multilinebox{\textbf{0.703} \\ \textbf{1023 sec}} & \multilinebox{\textbf{0.606} \\ \textbf{1939 sec}} & \multilinebox{0.842 \\ \textbf{3684 sec}} \\
\hline
\end{tabular}
\label{tab:num_samples}
\end{center}
\end{table}

\subsection{Performance Boost Using SAX}

\begin{figure}[ht!]
    \centering
    \setcounter{subfigure}{0}
    \subfloat[3 SAX segments.]{\includegraphics[width=0.4\textwidth]{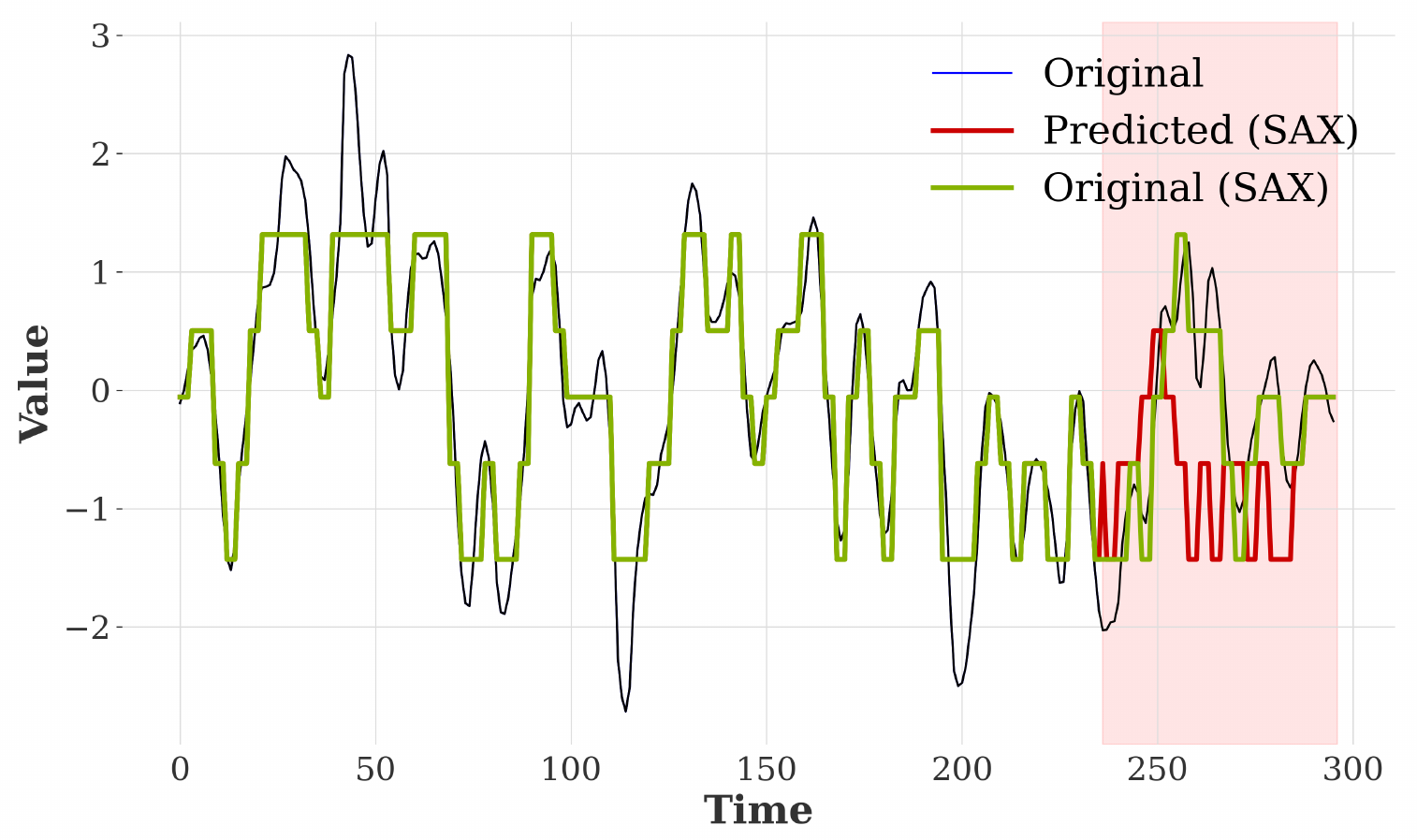}\label{subfig:3_seg}} \\
    \vspace{-10pt}
    \subfloat[6 SAX segments.]{\includegraphics[width=0.4\textwidth]{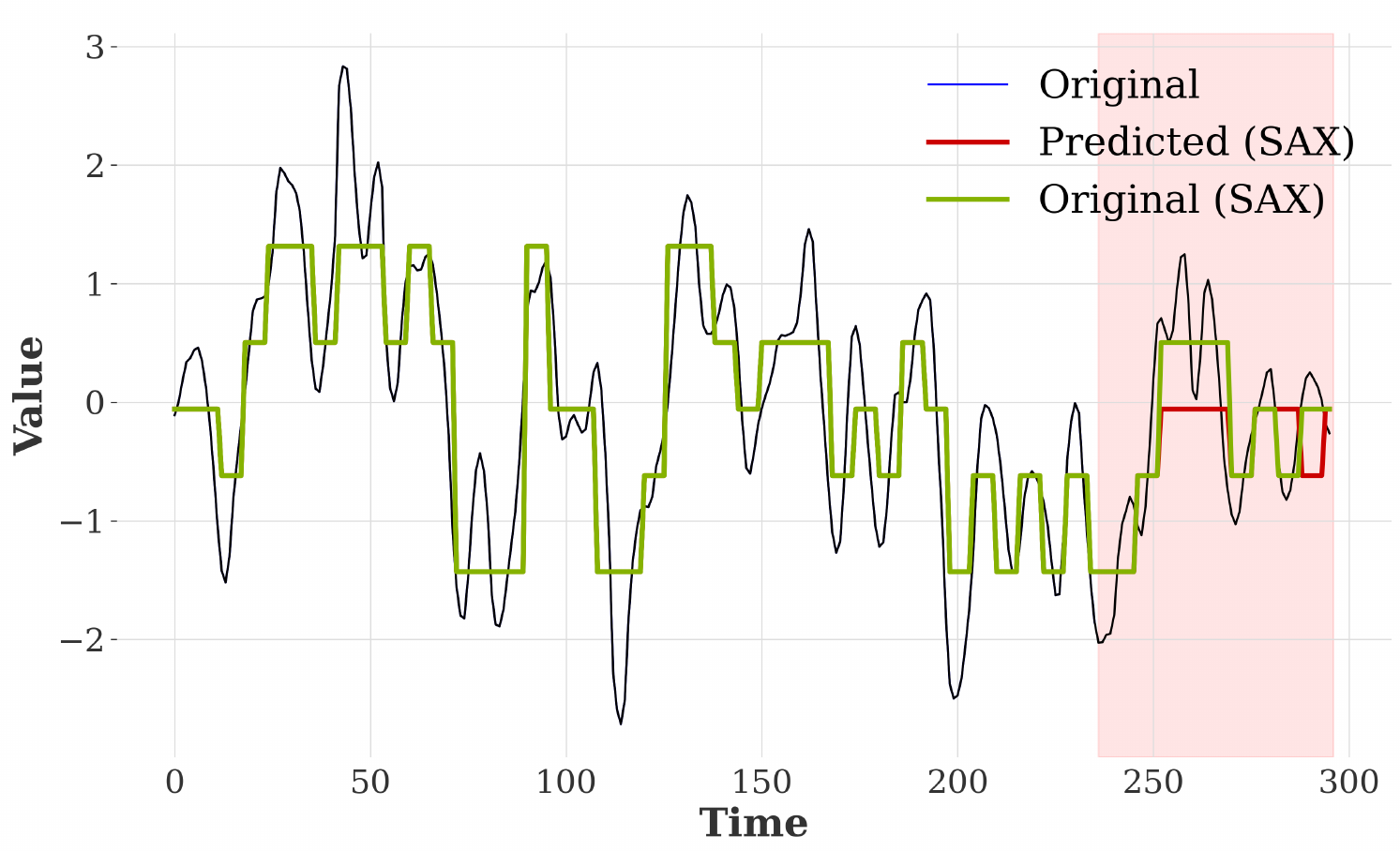}\label{subfig:6_seg}} \\
    \vspace{-10pt}
    \subfloat[9 SAX segments.]{\includegraphics[width=0.4\textwidth]{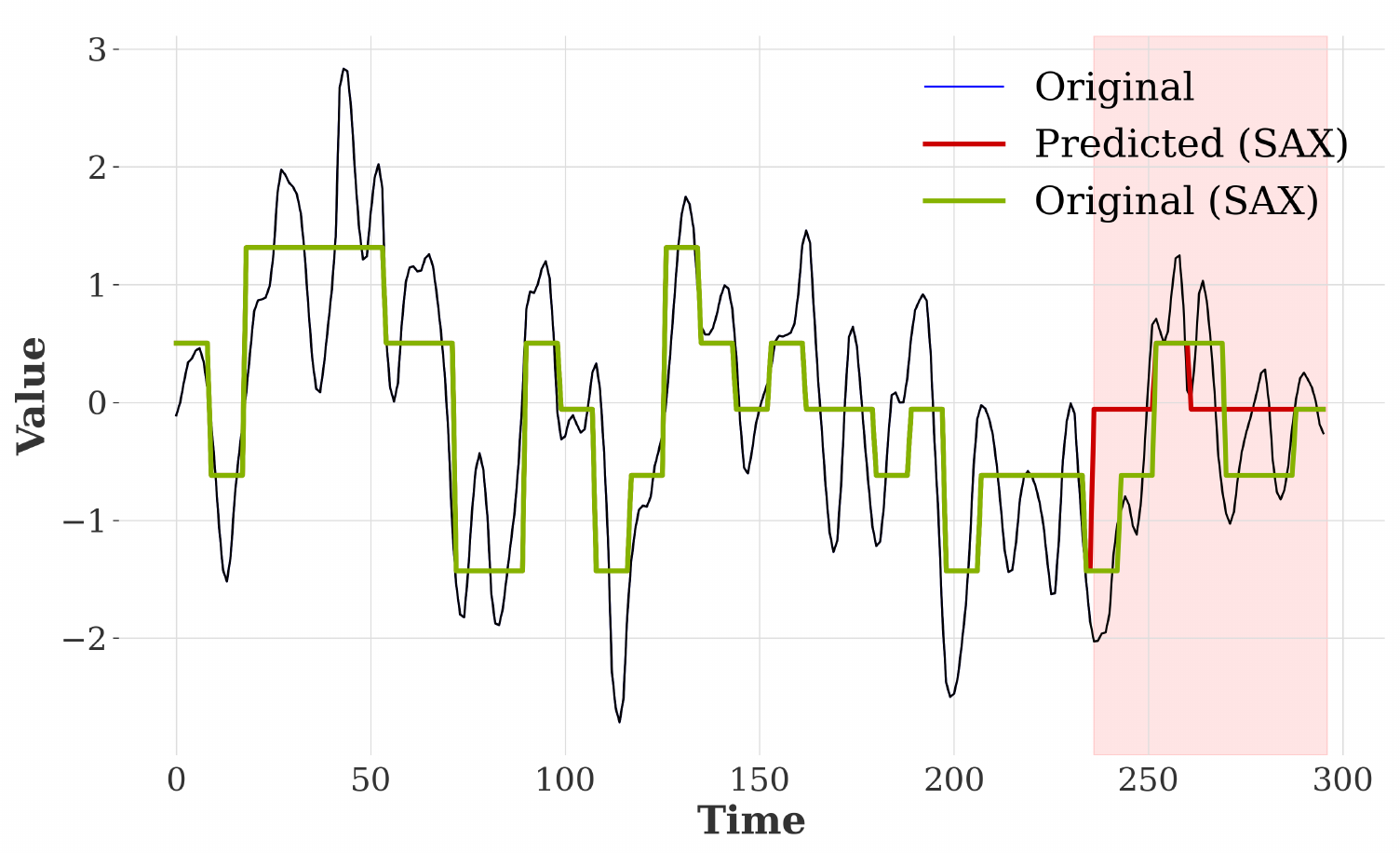}\label{subfig:9_seg}} \\
    \caption{Forecasting for various SAX segments (CO2\%).}
    \label{fig:sax_seg}
    \vspace{-6pt}
\end{figure}

\begin{figure}[ht!]
    \centering
    \setcounter{subfigure}{0}
    \subfloat[5 SAX symbols.]{\includegraphics[width=0.4\textwidth]{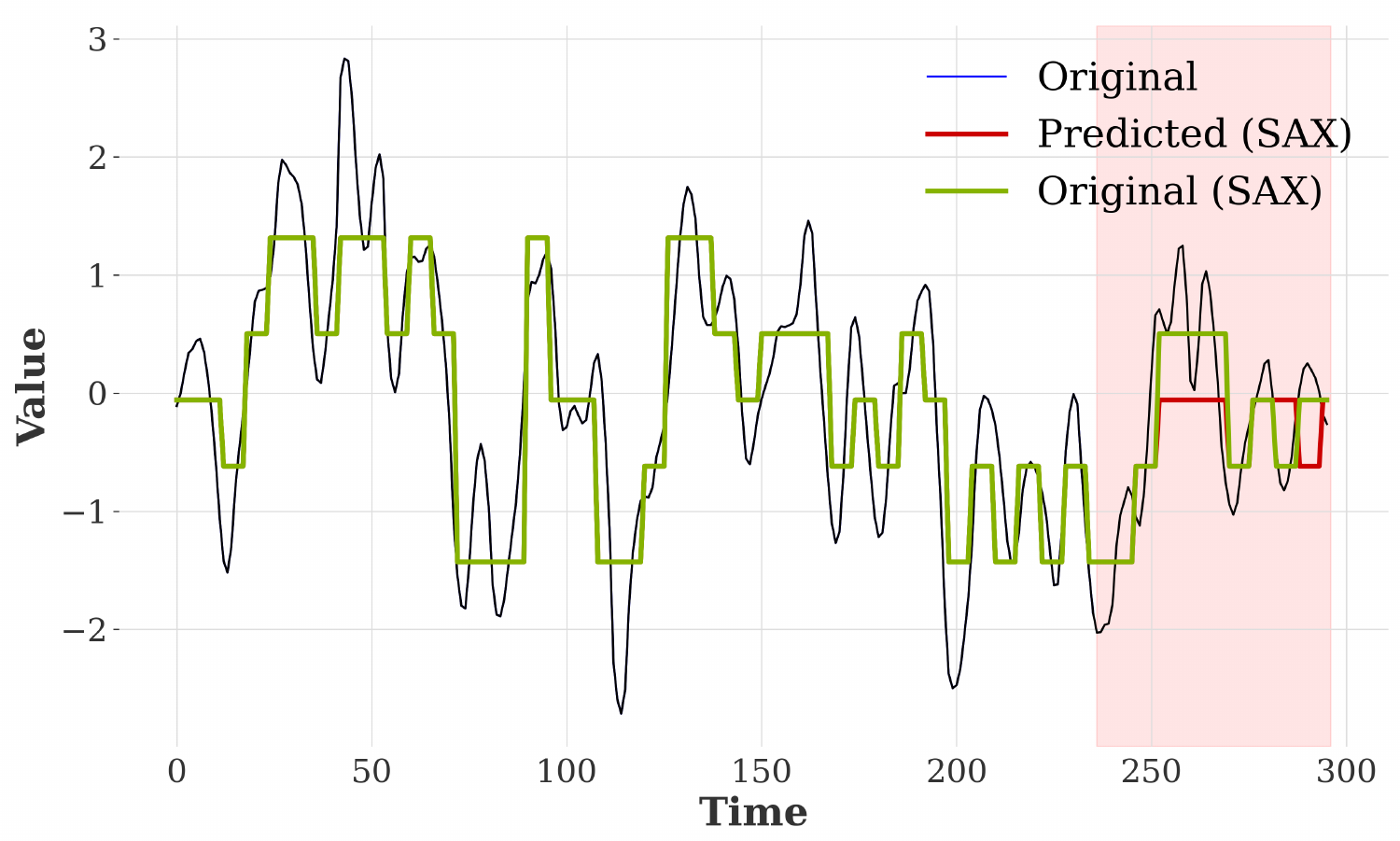}\label{subfig:5_symb}} \\
    \vspace{-10pt}
    \subfloat[10 SAX symbols.]{\includegraphics[width=0.4\textwidth]{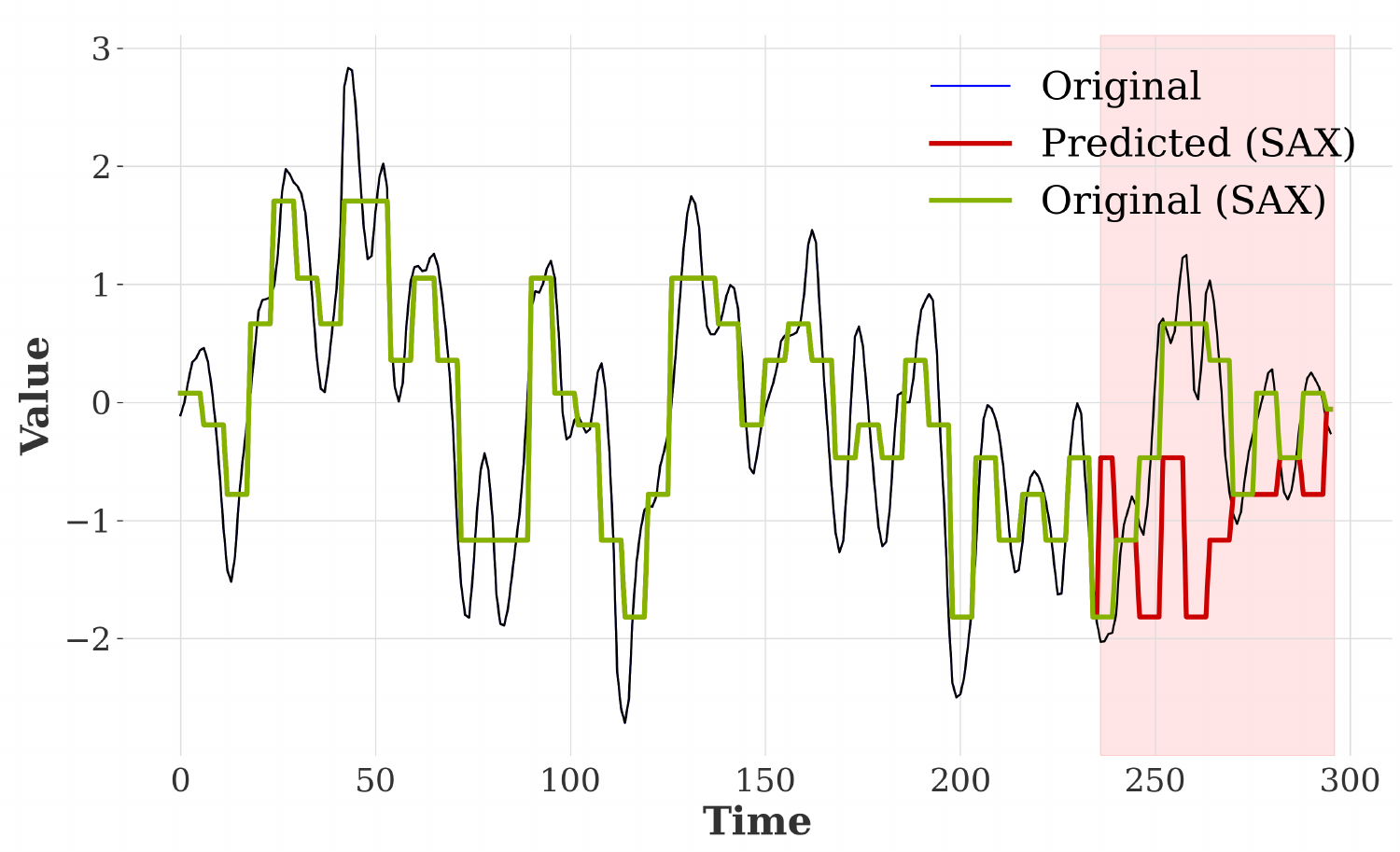}\label{subfig:10_symb}} \\
    \vspace{-10pt}
    \subfloat[20 SAX symbols.]{\includegraphics[width=0.4\textwidth]{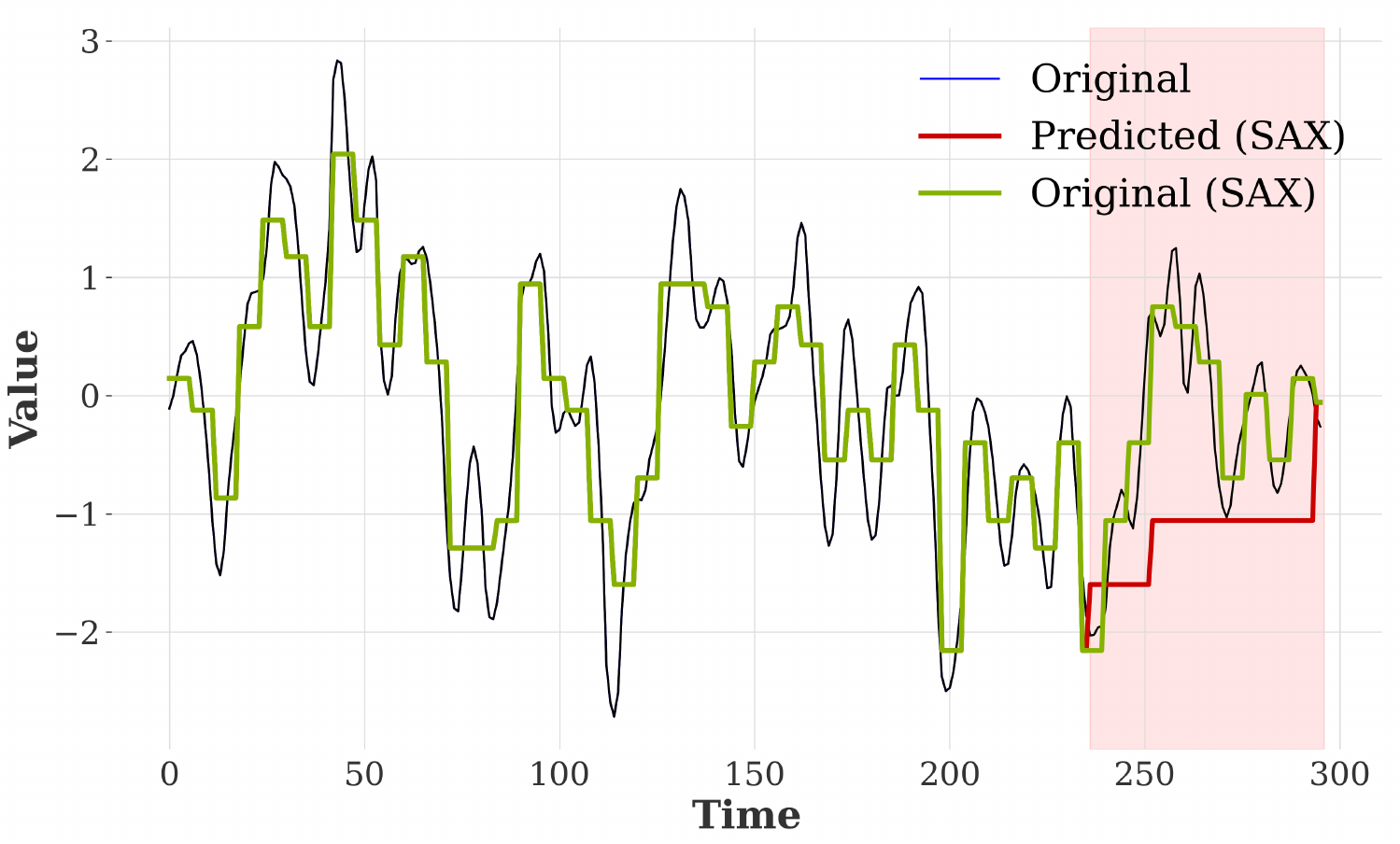}\label{subfig:20_symb}} \\
    \caption{Forecasting for different SAX alphabet sizes (CO2\%).}
    \label{fig:alph_size}
    \vspace{-6pt}
\end{figure}

\begin{figure}[ht!]
    \centering
    \includegraphics[width=0.4\textwidth]{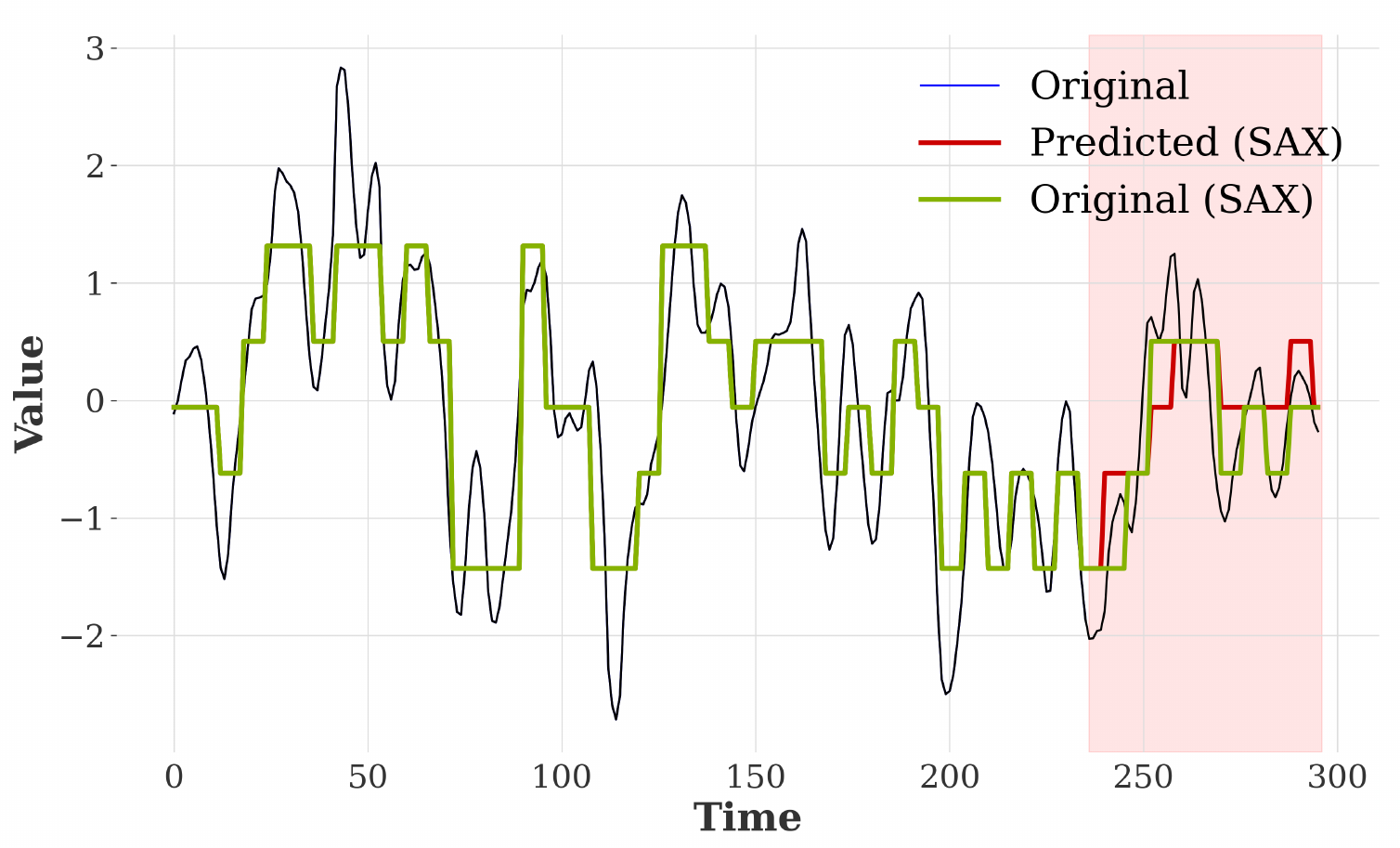}
    \vspace{-10pt}
    \caption{Forecasting using digits instead of letters as symbols.}
    \label{fig:digits}
\end{figure}

\label{subsec:perf_sax}
 Next, we will show the results obtained when quantization is applied using the SAX method, as described in Section~\ref{subsec:dim_mult}. Specifically, we evaluate the effects of increasing the length of the SAX segment and the size of the alphabet on the performance of zero-shot time series forecasting using MultiCast.

\subsubsection{Increasing SAX Segment length}
Table~\ref{tab:sax_length} lists the results for an increasing number of SAX segments for the CO2$\%$ dimension of the Gas Rate dataset, in terms of RMSE and execution time. We also list the results for using a different kind of SAX quantization; either using alphabetical characters or digits to encode SAX words. Compressing the time series significantly facilitates the inference process since now the model has to generate only one symbol per timestamp, instead of three or more. This is reflected in the execution times shown in Table~\ref{tab:sax_length}; inference after applying SAX compression is more than an order of magnitude faster, from 52 seconds in the best case (i.e., using 9 SAX segments) to 1168 seconds, when no quantization is applied. The large difference in performance can have a big impact on forecasting tasks that are run on CPU, which may often be the case in scenarios where access to a GPU with large enough memory to fit an LLM is not possible.


\begin{table}[ht!]
\caption{Increasing SAX segment length.}
\vspace{-12pt}
\begin{center}
\scriptsize
\renewcommand{\arraystretch}{1.2}
\begin{tabular}{|c|c|c|c|}
\hline
\multirow{2}{0.8cm}{\textbf{Method}} & \multicolumn{3}{c|}{\textbf{SAX Segment Length}} \\
\cline{2-4}
& \textbf{3} & \textbf{6} & \textbf{9} \\
\hline
MultiCast SAX (alphabetical) & \multilinebox{1.089 \\ 148 sec} & \multilinebox{0.983 \\ 77 sec} & \multilinebox{0.888 \\ 54 sec} \\
MultiCast SAX (digital) & \multilinebox{0.992 \\ 156 sec} & \multilinebox{0.99 \\ 71 sec} & \multilinebox{0.912 \\ \textbf{52 sec}}  \\
MultiCast & \multicolumn{3}{c|}{\multilinebox{\textbf{0.781} \\ 1168 sec }}  \\
\hline
\end{tabular}
\label{tab:sax_length}
\end{center}
\vspace{-15pt}
\end{table}

As expected, quantizing the time series leads to a loss of information. Again, this is reflected in the RMSE scores for the SAX approaches in Table~\ref{tab:sax_length}, which are worse than when no quantization is applied. However, this may not always be the case; having to infer only one symbol per timestamp is easier for the LLM. Patterns, if they exist, will be easier to detect and guess. The higher RMSE scores in these cases can be attributed to the quantization that SAX applies on both axes. However, the final result, when plotted, could properly follow the initial time series. This effect is illustrated in Figures~\ref{subfig:6_seg} and~\ref{subfig:9_seg}, for the CO2$\%$ dimension of the Gas Rate dataset. On the other hand, in this case, MultiCast using 3 SAX segments managed to detect the initial upward trend (Figure~\ref{subfig:3_seg}), but the result worsened afterwards.

Figure~\ref{fig:digits} shows an indicative example of the prediction result when applying SAX quantization using digits to encode symbols, for the CO2$\%$ dimension of the Gas Rate dataset. It is easily noticeable that the resulting line (red in the figure) closely follows the initial time series in this dimension. This could suggest that it may be easier for the LLMs to detect patterns in time series represented by numbers rather than alphabetical characters.

\subsubsection{Increasing SAX Alphabet Size}
In the following, we evaluate the performance of MultiCast, in terms of RMSE and execution time, when increasing the size of the SAX alphabet. Table~\ref{tab:sax_alph_size} lists the results. We should note that for digits we can only go up to an alphabet of size 10. Again, the non-quantized MultiCast yields better performance in terms of RMSE, but is significantly slower. Increasing the size of the alphabet does not seem to affect the execution time. Also, in terms of RMSE, larger alphabet sizes produced higher errors, possibly due to the increase in complexity that the use of more symbols induces.

\begin{table}[ht!]
\caption{Increasing SAX alphabet size.}
\vspace{-12pt}
\begin{center}
\scriptsize
\renewcommand{\arraystretch}{1.2}
\begin{tabular}{|c|c|c|c|}
\hline
\multirow{2}{0.8cm}{\textbf{Method}} & \multicolumn{3}{c|}{\textbf{SAX Alphabet Size}} \\
\cline{2-4}
& \textbf{5} & \textbf{10} & \textbf{20} \\
\hline
MultiCast SAX (alphabetical) & \multilinebox{0.983 \\ 77 sec} & \multilinebox{1.198 \\ 81 sec} & \multilinebox{1.273 \\ 83 sec} \\
MultiCast SAX (digital) & \multilinebox{0.99 \\ \textbf{71 sec}} & \multilinebox{1.21 \\ 75 sec} & N/A  \\
MultiCast & \multicolumn{3}{c|}{\multilinebox{\textbf{0.781} \\ 1168 sec }}  \\
\hline
\end{tabular}
\label{tab:sax_alph_size}
\end{center}
\end{table}

Finally, Figure~\ref{fig:alph_size} shows an indicative forecasting example for 5, 10 and 20 SAX symbols for the CO2$\%$ dimension of the Gas Rate dataset. The drop in RMSE scores is also reflected here; only in the case of using five symbols does the forecasted time series follow the trend of the original.
\section{Conclusions}
\label{sec:conclusions}

In this paper, we presented MultiCast, an approach that leverages LLMs for zero-shot multivariate time series forecasting. To make this model work with multiple dimensions, we proposed three token multiplexing solutions that reduce the dimensionality of the time series to one. This allows the time series to be compatible with the input of an LLM, while still preserving its ability to detect repetitive patterns. Additionally, we presented a quantization solution that aims to facilitate the learning of existing patterns in the series by LLMs. This solution also significantly reduces the execution time. In our comprehensive experimental analysis using three real-world datasets, we found that the use of LLMs for multivariate zero-shot time series forecasting shows promise and offers a significant advantage compared to other similar methods available in the literature: \textit{No expert knowledge or time and resource-consuming parameter search processes are required}. In the future, we plan to expand our research on employing LLMs for zero-shot solutions on other similar time series-related tasks, such as imputation, anomaly detection, and change point detection, as well as evaluate MultiCast's inference performance using more LLMs as back-end models and further improving it in more dimensions.

\section*{Acknowledgment}
This work was partially funded by the EU Horizon Europe projects STELAR (101070122) and DT4GS (101056799).

\bibliographystyle{IEEEtran}
\bibliography{References}

\end{document}